\documentclass[conference]{IEEEtran}
\IEEEoverridecommandlockouts
\usepackage{cite}
\usepackage{amsmath,amssymb,amsfonts}
\usepackage{algorithm}
\usepackage{algpseudocode}

\usepackage{graphicx}
\usepackage{textcomp}
\usepackage{xcolor}
\usepackage{soul}
\usepackage{scalerel}
\usepackage{stackengine}
\usepackage{multirow}

\usepackage{color}

\def\BibTeX{{\rm B\kern-.05em{\sc i\kern-.025em b}\kern-.08em
    T\kern-.1667em\lower.7ex\hbox{E}\kern-.125emX}}

\usepackage[colorlinks=true,linkcolor=red,citecolor=blue, urlcolor=blue]{hyperref}
\usepackage{tabularray}

\newcommand{\dsquare}{\hspace*{2pt}\topinset{$\square$}{$\square$}{2pt}{-2pt}}
\newcommand{\ssquare}{\topinset{$\square$}{$\square$}{0.1pt}{-0.1pt}}

\newcommand\dunderline[3][-1pt]{{%
  \sbox0{#3}%
  \ooalign{\copy0\cr\rule[\dimexpr#1-#2\relax]{\wd0}{#2}}}}
\def\subsubsec#1{\noindent\scaleto{\dunderline[-.75pt]{0.25pt}{\textbf{#1}}.}{7.6pt}}

\newlength{\blob}
\def\ptext#1{\color{green!60!black!90} {\scaleto{(#1)}{6.5pt}} \color{black}}
\def\ntext#1{\color{red} {\scaleto{(#1)}{6.5pt}} \color{black}}
\def\sdtext#1{\color{black} {\scriptsize $\pm$ \hspace{-1.4mm} #1} \color{black}}

\def \name {Spade}

\title{Keep Your Friends Close \& Enemies Farther: \\
Debiasing Contrastive Learning with Spatial Priors in 3D Radiology Images
\thanks{978-1-6654-6819-0/22/\$31.00 ©2022 IEEE}
}

\makeatletter
\newcommand{\linebreakand}{%
  \end{@IEEEauthorhalign}
  \vspace{-3mm}
  \hfill\mbox{}\par
  \mbox{}\hfill\begin{@IEEEauthorhalign}
}
\makeatother

\author{
\IEEEauthorblockN{Yejia Zhang}
\IEEEauthorblockA{\textit{University of Notre Dame} \\
                  Notre Dame, IN, 46556, USA \\
                  yzhang46@nd.edu} 
\and
\IEEEauthorblockN{Nishchal Sapkota}
\IEEEauthorblockA{\textit{University of Notre Dame} \\
                  Notre Dame, IN, 46556, USA \\
                  nsapkota@nd.edu}
\and
\IEEEauthorblockN{Pengfei Gu}
\IEEEauthorblockA{\textit{University of Notre Dame} \\
                  Notre Dame, IN, 46556, USA \\
                  pgu@nd.edu}
\linebreakand
\IEEEauthorblockN{Yaopeng Peng}
\IEEEauthorblockA{\textit{University of Notre Dame} \\
                  Notre Dame, IN, 46556, USA \\
                  ypeng4@nd.edu}
\and
\IEEEauthorblockN{Hao Zheng}
\IEEEauthorblockA{\textit{University of Notre Dame} \\
                  Notre Dame, IN, 46556, USA \\
                  hzheng3@nd.edu}
\and
\IEEEauthorblockN{Danny Z. Chen}
\IEEEauthorblockA{\textit{University of Notre Dame} \\
                  Notre Dame, IN, 46556, USA \\
                  dchen@nd.edu}
}

\makeatletter
\patchcmd{\@maketitle}
  {\addvspace{0.5\baselineskip}\egroup}
  {\addvspace{-1\baselineskip}\egroup}
  {}
  {}
\makeatother

\begin{document}




\maketitle

\begin{abstract}
Understanding of spatial attributes is central to effective 3D radiology image analysis where crop-based learning is the de facto standard.
Given an image patch, its core spatial properties (e.g., position \& orientation) provide helpful priors on expected object sizes, appearances, and structures through inherent anatomical consistencies.
Spatial correspondences, in particular, can effectively gauge semantic similarities between inter-image regions, while their approximate extraction requires no annotations or overbearing computational costs.
However, recent 3D contrastive learning approaches either neglect correspondences or fail to maximally capitalize on them.
To this end, we propose an extensible 3D contrastive framework (\name, for \underline{Spa}tial \underline{De}biasing) that leverages extracted correspondences to select more effective positive \& negative samples for representation learning.
Our method learns both globally invariant and locally equivariant representations with downstream segmentation in mind. 
We also propose separate selection strategies for global \& local scopes that tailor to their respective representational requirements. 
Compared to recent state-of-the-art approaches, Spade shows notable improvements on three downstream segmentation tasks (CT Abdominal Organ, CT Heart, MR Heart).

\end{abstract}




\section{Introduction} \label{intro}

Deep neural networks have the ability to learn useful features with the caveat of requiring large amounts of annotated training data.
To reduce the expense of obtaining sample labels, self-supervised contrastive learning in natural scene images \cite{Chen2020ImprovedBW}\cite{Oord2018RepresentationLW}\cite{Wu2018UnsupervisedFLNonparam}\cite{chen2020simclr}\cite{chuang2020debiasedcon} use instance discrimination with the objective of making latent representations of similar images (i.e., \textit{positives} or augmented views of the same image) close while contrasting features of distinct images (i.e., \textit{negatives} or views from other images).

Many differing characteristics of radiology images, however, make direct applications of this idea suboptimal.
Instance discrimination is especially problematic due to the existence of recurring anatomical structures in volumes which often produces ``negatives'' with similar semantics (i.e., \textit{false negatives}).
Presence of false negatives slows convergence, impairs representations, and discards useful semantics \cite{huynh2022boosting}.
These methods also favor global, invariant features whereas many medical localization tasks benefit from local, equivariant representations.
Addressing false negatives, recent methods propose improved selections of positives by 
utilizing subject IDs \cite{azizi2021bigmedical}, patient metadata \cite{Sowrirajan2021MoCoPIMetadata}, and spatial correspondences among image slices \cite{Zeng2021PositionalCL}.
These kill two birds with one stone by both removing false negatives and expanding the diversity of positives.
For better local representations, \cite{chaitanya2020contrastive} also applies instance discrimination on image sub-crops on top of global contrastive learning which also selects positives based on slice positions. 


Despite these advances, three important factors conducive to learning representations for 3D radiology image segmentation remain unaddressed.
\textbf{(1)} The use of 2D slices discards spatial context and makes pretrained models incompatible with volumetric fine-tuning methods.
Recent works affirm this by showing that pretraining \cite{Zhou2019ModelsGG} and fine-tuning \cite{isensee2021nnu} 3D tasks on 3D data outperform 2D equivalents. 
Plus, using whole slices introduces multiple anatomical structures which may further impair representations by inviting short-cut learning where models over-focus on discriminative regions and neglect remaining objects. 
\textbf{(2)} The benefits of contrastive learning are not leveraged in decoder outputs where discriminative representations are more important for downstream localization tasks.
Although this was partially addressed in~\cite{chaitanya2020contrastive}, the final method only uses local samples from the same image, which does not fully capitalize on spatial priors that facilitate learning of anatomical patterns across images.
This brings us to our final point.
\textbf{(3)} Both the diversity \& quality of positive \& negative sample candidates are limited either because samples are constrained to small selections of intra-batch images \cite{chaitanya2020contrastive}\cite{Zeng2021PositionalCL}\cite{you2022simcvd} or that inter-image positives are precluded altogether \cite{Zhou2021PreservationalLI}\cite{Zheng2021HierarchicalSL}.
Medical images exhibit less appearance variation than natural scene images which accentuates the importance of inter-image sample diversity for positives \& negatives. 

\begin{figure*}[!t]
    \centering
    \includegraphics[width=0.6\linewidth]{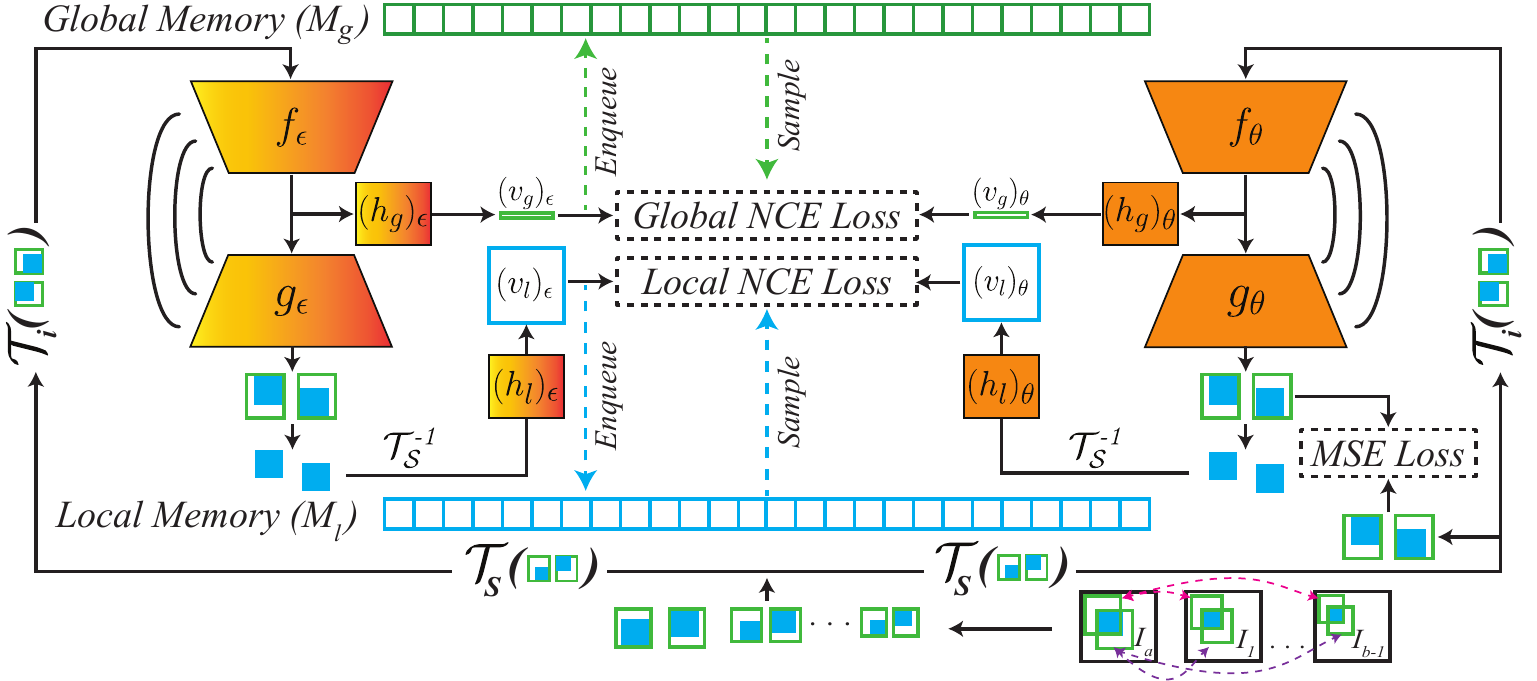}
    \vspace*{-4mm}
    \caption{The overall workflow of the Spade framework. $\mathcal{T}_s$ \& $\mathcal{T}_i$ represent spatial \& intensity transforms, respectively.}
    \label{fig:framework}
\vspace*{-5mm}
\end{figure*}

In this work, we address these drawbacks with a new 3D self-supervised framework that builds on MoCo \cite{Chen2020ImprovedBW} (i.e., applying contrastive learning on encoder features with a global memory bank) and MG \cite{Zhou2019ModelsGG} (i.e., employing reconstruction on decoder outputs) to pretrain a UNet-like model.
We utilize 3D crops (ancillary to points made in \textbf{(1)}) since cropping naturally regularizes features by restricting image contents and promotes more equitable foci among anatomical structures.
For \textbf{(2)}, 
we propose a local contrastive component with a separate local memory bank that boosts features from the decoder
and learns spatially equivariant representations which benefit downstream segmentation tasks more than solely promoting global, invariant features (e.g., \cite{Zhou2021PreservationalLI}\cite{Zheng2021HierarchicalSL}).
Finally, to address \textbf{(3)}, we go beyond instance discrimination and leverage spatial priors (i.e., correspondences) in crops across volumes to select higher quality positives \textit{and} negatives. 
Here, we emphasize correspondences for three primary reasons: they are \textit{free} (i.e., require no annotations to obtain), \textit{tractable} (i.e., involve little computation), and \textit{effective} (i.e., disclose helpful information for gauging anatomical similarity).
Thus, correspondences between images are first computed by aligning them to a template image (details in \autoref{method}) since suitable templates are accessible (e.g., selecting a representative image within the dataset) and computing alignment is cheap (1.8 seconds per CT volume). 
With correspondences as a proxy for semantic similarity, we propose multiple global \& local sampling strategies to reduce false negatives (i.e., debiasing) and increase the diversity \& quality of positives.
Our final framework, \textbf{Spade} (for \textbf{Spa}tial \textbf{de}biasing), incorporates the best local \& global strategies.

To demonstrate Spade's efficacy (see \autoref{res}), we utilize 3D torso radiology images.
More specifically, we pretrain using 623 chest CT \& 200 abdominal CT images, and fine-tune on three downstream segmentation tasks (CT Abdominal, CT Heart, MR Heart). 
We conduct thorough studies on Spade's components and the proposed global \& local sampling strategies to explore how spatial priors facilitate contrastive learning in 3D radiology images. 
Additionally, we study Spade's robustness to template choice and show that rough alignment is both cheap and effective for downstream segmentation.
Finally, we compare Spade to recent self-supervised methods and report sizable performance improvements over state-of-the-art predictive, reconstructive, and contrastive pretraining methods (e.g., +2.7\%, +3.4\%, +1.8\% dice, resp., on CT Heart).

\section{Methodology} \label{method}

Before detailing Spade's workflow \& components (see Fig.~\ref{fig:framework}), we describe template selection \& contrastive learning.

\subsection{Preliminaries}



\noindent\textbf{Template Selection \& Spatial Correspondence}. \label{method1a}
The main objective of aligning images to a chosen template is to establish spatial correspondences through a shared coordinate system in which we can better gauge semantic similarity. 
Given an unlabeled dataset with $N$ images, $D$=$\{I_1, I_2, \ldots ,I_N\}$, we define a template image as $I_t$ (note that $I_t \in D$ is valid) such that the template contains all key anatomical structures present in $D$.
We then register all the images in $D$ to $I_t$ by computing $\mathbb{T}$=$\{T_1, T_2, \cdots ,T_N\}$, where $T_i: (x_i, y_i, z_i) \rightarrow (x^t_i, y^t_i, z^t_i)$ is a function that transforms $I_i$'s coordinates to $I_t$'s. 
We apply affine registration and optimize with SGD and cross-correlation as the similarity metric.

Thus, given a patch $P_i$ from image $I_i$, we find its corresponding crop in $I_t$ with $P_t$=$T_i(P_i)$.
Further, $P_i$'s corresponding crop in any other image $I_j\in D$ can be extracted via $P_j$=$T_j^{-1} \circ T_i(P_i)$.
We concisely denote the corresponding patch of $P_i$ in image $I_j$ as $P_{i \rightarrow j}$.
Note that we do not transform the images, rather, we only compute $\mathbb{T}$.
\\
\noindent\textbf{Contrastive Representation Learning}. \label{method1b}
We denote the $L_2$-normalized unit embedding of an anchor patch as $v$.
Borrowing notation from \cite{chaitanya2020contrastive}, we define a positive sample as $v^+ \in \Lambda^+$, where $\Lambda^+$ is the set of positive embedding vectors for anchor $v$.
Similarly, a negative sample is defined as $v^- \in \Lambda^-$, where $\Lambda^-$ is the set of negative embedding vectors.
The contrastive loss (NCE \cite{Oord2018RepresentationLW} for short) is then defined as:
\vspace{-2mm}
\begin{equation} \label{nce}
\vspace{-2mm}
{\footnotesize{\mathcal{L}^{NCE} = - \log \frac{\exp({v \cdot v^+ / \tau)}}
                            {\exp({v \cdot v^+ / \tau)} + \sum\limits_{v^-\in \Lambda^-_{g/l}} \exp(v \cdot v^- / \tau)}}}
\end{equation}
where ``$\mathcal{L}^{NCE}$'' is shorthand for the loss $\mathcal{L}^{NCE}(v, v^+)$ between an anchor $v$ and a positive $v^+$, $\tau$ is a temperature hyperparameter, and $g$ \& $l$ indicate global \& local components respectively.
Incorporating all positive samples, the complete contrastive loss (CON for short) is written as:
\vspace{-2mm}
\begin{equation} \label{con}
\vspace{-2mm}
{\footnotesize{\mathcal{L}^{CON}_{g/l} = \frac{1}{|\Lambda^+_{g/l}|} 
                            \sum\limits_{\forall(\dot{v}, \ddot{v}) \in \Lambda^+_{g/l}}
                            \mathcal{L}^{NCE}(\dot{v}, \ddot{v}) + 
                            \mathcal{L}^{NCE}(\ddot{v}, \dot{v})}}
\end{equation}


\subsection{The Spade Framework} \label{method2}

Spade pretrains a randomly initialized UNet-like network containing an encoder $f_\theta$, decoder $g_\theta$, global projection module $h_{\theta_g}$, local projection module $h_{\theta_l}$, and their corresponding momentum counterparts $f_\epsilon$, $g_\epsilon$, $h_{\epsilon_g}$, and $h_{\epsilon_l}$, respectively (see Fig.~\ref{fig:framework}).
For each batch, an anchor image $I_a$ and $p$ pairs of anchor patches $\{P_a^j$, $P_a^k, \ldots \}$ are sampled where overlaps of paired patches are at least $o$\% and $P \in \mathbb{R}^{1 \times D_{in} \times H_{in} \times W_{in}}$.
Following MoCo \cite{Chen2020ImprovedBW}, each $P$ is randomly transformed twice via spatial augmentations $\mathcal{T}_S$ (e.g., flipping, rotating) and intensity noising $\mathcal{T}_I$ (e.g., those proposed in \cite{Zhou2019ModelsGG}). 
All these views are inputted into both the regular \& momentum networks.

Spade contains three pretext tasks (i.e., global contrastive learning, local contrastive learning, and reconstruction). 
These tasks use global logits $z$=$f(P)$, and local logits $Z$=$g \circ f(P)$, where $f$ may refer to either $f_\theta$ or $f_\epsilon$ for brevity.
For global \& local contrastive learning, $v$ is a unit feature embedding obtained from the global projection module $v_g$=$h_g(z)$, $v_g\in\mathbb{R}^{C_g^{emb}}$ or the local projection module $v_l$=$h_l(Z)$, $v_l\in\mathbb{R}^{C_l^{emb} \times H_l^{emb} \times W_l^{emb}}$. 
Note that $v_g$ is a vector while $v_l$ has dimensionality $H_l^{emb} \times W_l^{emb}$ since we aim to preserve spatial equivariance for local representations.
After each iteration, global \& local embeddings from the momentum network are then enqueued into $M_g$ (size $\mathcal{Q}_g$) \& $M_l$ (size $\mathcal{Q}_l$), respectively, along with their corresponding positions in the template image.

\noindent\subsubsec{Global Feature Learning} \label{method2a} 
The objective of global feature learning is to learn high-level semantics from crops that are invariant to viewpoint changes or transformations.
Toward this end, we introduce several strategies to improve representations by further boosting the diversity \& efficacy of both positive and negative samples.
For pithiness, we formulate these from a single view of an anchor patch $P_a$ and its embedding $v_a$.

For \textbf{global strategy 1} (\textit{G1}), we augment positives with crops of corresponding positions from different images. 
This is similar in spirit to \cite{chaitanya2020contrastive}\cite{Zeng2021PositionalCL}, but we compute correspondences in 3D instead of assuming alignment and sample crops instead of slices which are both richer in spatial context \& more robust to alignment errors from a single axis.
Concretely, $\Lambda^-_{\textit{G1}}$=$M_g$, $\Lambda^+_{\textit{G1}}$=$\{v_a, v_{a \rightarrow i_1}, v_{a \rightarrow i_2}, \ldots, v_{a \rightarrow i_{n^+}}\}$, where $v_{a \rightarrow i}$=$h_g \circ f (P_{a \rightarrow i})$ denotes the feature embedding of $P_{a \rightarrow i}$, and $n^+$ is the number of sampled positives in separate volumes that spatially correspond to the anchor patch $P_a$. 

For a large embedding queue, quantity has a quality of its own, but naively raising this capacity increases incidences of false negatives.
In \textbf{global strategy 2} (\textit{G2}), we debias the negative cohort by removing crop embeddings in $M_g$ that are ``close'' to the anchor patch.
We quantify ``close'' as the overlap between two patches in the template space. 
Given two patches $P_a^j$ \& $P_a^k$, we denote their overlap region as patch $P_a^{j \cap k}$. 
Their overlap ratio or IoU is defined as $IoU(P_a^j, P_a^k)$=$V(P_a^{j \cap k}) / [V(P_a^j) + V(P_a^k) - V(P_a^{j \cap k})]$, where $V()$ returns the volume of a patch. 
Given an overlap threshold $o \in [0, 1]$, we define the cohorts of positives \& negatives as
$\Lambda^-_{\textit{G2}}$=$\{v_i \ | \ IoU(v_{i \rightarrow t}, v_{a \rightarrow t}) \leq o, \forall v_i \in M_g\}$ 
\& 
$\Lambda^+_{\textit{G2}}$=$\{v_a, v_{a \rightarrow i_1}, v_{a \rightarrow i_2}, \ldots, v_{a \rightarrow i_{n^+}}\}$, where $IoU(v_{i \rightarrow t}, v_{a \rightarrow t})$ is notational shorthand for $IoU(P_{i \rightarrow t}, P_{a \rightarrow t})$. 

Although false negatives are alleviated, $\Lambda^+_{\textit{G1}}$ and $\Lambda^+_{\textit{G2}}$ still suffer from the same drawbacks as \cite{Zeng2021PositionalCL}\cite{chaitanya2020contrastive}. 
Reliance on intra-batch samples greatly limits the variability of positives just as it had for negatives.
In light of this, we propose \textbf{global strategy 3} (\textit{G3}) which converts the removed negative patches in \textit{G2} to positives based on the reasoning that if the semantics of a patch are similar enough to be considered a false negative, then the mutual information present would enrich the representations with added positives. 
We define
$\Lambda^-_{\textit{G3}}$=$\{v_i \ | \ IoU(v_{i \rightarrow t}, v_{a \rightarrow t}) \leq o, \forall v_i \in M_g\}$ 
\& 
$\Lambda^+_{\textit{G3}}$=$\{v_a, v_{a \rightarrow i_1}, v_{a \rightarrow i_2}, \ldots, v_{a \rightarrow i_n^+}\} \cup \{v_i \ | \ IoU(v_{i \rightarrow t}, v_{a \rightarrow t}) > o, \forall v_i \in M_g\}$.



\begin{table*}[!h]
\begin{center}
\caption{\label{tab:main}\textbf{Main results with state-of-the-art approaches.} 
         Entries are dice scores (\%) and their standard deviations ($\pm$) across 4 runs. }
\vspace*{-3mm}
\resizebox{0.85\textwidth}{!}{
\begin{tblr}{
    columns={colsep=3pt},
    colspec={l || l l l | l l l | l l l },
    row{1-3} = {gray!50!black!15},
    row{5,10} = {gray!20!black!5}
    }
\hline 
\multirow{3}{*}{\hspace*{5.4mm} Methods} & \multicolumn{3}{c} {CT Abdominal (BCV)} & \multicolumn{3}{c} {CT Heart (MMWHS)} & \multicolumn{3}{c} {MR Heart (MMWHS)} \\
\cline{2-10} 
    & 10\% & 25\% & 50\%  & 10\% & 25\% & 50\% & 10\% & 25\% & 50\% \\
    & 2 & 5 & 10 & 1 & 3 & 7 & 1 &3 &7 \\
\hline \hline
\hspace{-1mm}\makebox[\blob][l]{ } Random Init. & 
52.72 \hspace{-2pt}\sdtext{1.03} & 69.37 \hspace{-2pt}\sdtext{0.79} & 78.60 \hspace{-2pt}\sdtext{0.40} 
& 62.49 \hspace{-2pt}\sdtext{1.18} & 85.23 \hspace{-2pt}\sdtext{0.68} & 89.01 \hspace{-2pt}\sdtext{0.51} & 
68.45 \hspace{-2pt}\sdtext{1.15} & 75.70 \hspace{-2pt}\sdtext{3.27} & 86.26 \hspace{-2pt}\sdtext{0.68} \\ [-0.2ex] 
\hline 
\multicolumn{10}{l}{\hspace*{0.9mm}\textit{Predictive \& Generative Approaches}} \\ [-0.2ex] 
\hline
\hspace{-1mm}\makebox[\blob][r]{\cite{Zhou2019ModelsGG}} MG{\tiny{$_{19}$}} & 
54.10 \hspace{-2pt}\sdtext{1.45} & \underline{71.78} \hspace{-2pt}\sdtext{0.84} & 79.39 \hspace{-2pt}\sdtext{0.71} & 
64.02 \hspace{-2pt}\sdtext{1.59} & 86.20 \hspace{-2pt}\sdtext{1.27} & 89.08 \hspace{-2pt}\sdtext{0.25} & 
69.66 \hspace{-2pt}\sdtext{1.47} & 74.88 \hspace{-2pt}\sdtext{3.99} & 86.01 \hspace{-2pt}\sdtext{0.30} \\
\hspace{-1mm}\makebox[\blob][r]{\cite{tao2020revisiting}} Rubik++{\tiny{$_{20}$}} & 
54.99 \hspace{-2pt}\sdtext{1.45} & 69.75 \hspace{-2pt}\sdtext{0.56} & 79.80 \hspace{-2pt}\sdtext{0.46} & 
64.67 \hspace{-2pt}\sdtext{3.25} & 86.31 \hspace{-2pt}\sdtext{0.74} & 89.02 \hspace{-2pt}\sdtext{0.17} & 
72.49 \hspace{-2pt}\sdtext{1.27} & 78.31 \hspace{-2pt}\sdtext{1.42} & 86.86
\hspace{-2pt}\sdtext{0.32} \\
\hspace{-1mm}\makebox[\blob][r]{\cite{Zhang2021SARSR}} SAR{\tiny{$_{21}$}} & 
53.14 \hspace{-2pt}\sdtext{1.68} & 69.96 \hspace{-2pt}\sdtext{0.38} & 78.31 \hspace{-2pt}\sdtext{0.50} & 
65.01 \hspace{-2pt}\sdtext{1.49} & 86.50 \hspace{-2pt}\sdtext{0.88} & 89.18 \hspace{-2pt}\sdtext{0.35} & 
73.25 \hspace{-2pt}\sdtext{1.45} & \underline{78.60} \hspace{-2pt}\sdtext{1.34} & 86.72 \hspace{-2pt}\sdtext{0.13} \\
\hspace{-1mm}\makebox[\blob][r]{\cite{zHaghighi2021TransferableVW}} TransVW{\tiny{$_{21}$}} & 
55.42 \hspace{-2pt}\sdtext{1.86} & 71.58 \hspace{-2pt}\sdtext{1.33} & 79.66 \hspace{-2pt}\sdtext{1.07} & 
65.21 \hspace{-2pt}\sdtext{1.27} & 86.48 \hspace{-2pt}\sdtext{1.23} & \underline{90.17} \hspace{-2pt}\sdtext{0.45} & 
71.91 \hspace{-2pt}\sdtext{2.08} & 77.32 \hspace{-2pt}\sdtext{1.56} & 86.30 \hspace{-2pt}\sdtext{0.57}\\
\hline
\multicolumn{10}{l}{\hspace*{0.9mm}\textit{Metric Learning Approaches}} \\ [-0.2ex] 
\hline
\hspace{-1mm}\makebox[\blob][r]{\cite{Xie2020PGLPL}} PGL{\tiny{$_{20}$}} & 
54.87 \hspace{-2pt}\sdtext{1.12} & 70.45 \hspace{-2pt}\sdtext{0.98} & 78.88 \hspace{-2pt}\sdtext{0.27} & 
63.68 \hspace{-2pt}\sdtext{2.85} & 86.65 \hspace{-2pt}\sdtext{0.77} & 88.84 \hspace{-2pt}\sdtext{0.41} & 
68.61 \hspace{-2pt}\sdtext{2.86} & 74.29 \hspace{-2pt}\sdtext{3.55} & 86.56 \hspace{-2pt}\sdtext{0.68}\\
\hspace{-1mm}\makebox[\blob][r]{\cite{Chen2020ImprovedBW}} MoCo{\tiny{$_{20}$}} & 
55.64 \hspace{-2pt}\sdtext{0.85} & 71.07 \hspace{-2pt}\sdtext{1.07} & 79.97 \hspace{-2pt}\sdtext{0.66} & 
65.03 \hspace{-2pt}\sdtext{1.62} & 85.96 \hspace{-2pt}\sdtext{1.21} & 89.63 \hspace{-2pt}\sdtext{0.34} & 
69.44 \hspace{-2pt}\sdtext{0.80} & 77.23 \hspace{-2pt}\sdtext{1.60} & 86.51 \hspace{-2pt}\sdtext{0.13}\\
\hspace{-1mm}\makebox[\blob][r]{\cite{Zeng2021PositionalCL}} PCL{\tiny{$_{21}$}} & 
\underline{56.05} \hspace{-2pt}\sdtext{0.64} & 68.55 \hspace{-2pt}\sdtext{0.55} & 76.11 \hspace{-2pt}\sdtext{0.42} & 
\underline{66.23} \hspace{-2pt}\sdtext{1.25} & 85.14 \hspace{-2pt}\sdtext{1.01} & 88.25 \hspace{-2pt}\sdtext{0.32} & 
73.44 \hspace{-2pt}\sdtext{1.08} & 76.90 \hspace{-2pt}\sdtext{1.05} & 83.50 \hspace{-2pt}\sdtext{0.54} \\
\hspace{-1mm}\makebox[\blob][r]{\cite{Zhou2021PreservationalLI}} PCRL{\tiny{$_{21}$}} & 
56.01 \hspace{-2pt}\sdtext{1.39} & 71.30 \hspace{-2pt}\sdtext{0.91} & \textbf{80.23} \hspace{-2pt}\sdtext{0.24} & 
65.58 \hspace{-2pt}\sdtext{2.41} & \underline{87.03} \hspace{-2pt}\sdtext{0.98} & 90.02 \hspace{-2pt}\sdtext{0.70} & 
\underline{74.83} \hspace{-2pt}\sdtext{1.91} & 77.72 \hspace{-2pt}\sdtext{1.97} & \underline{86.79} \hspace{-2pt}\sdtext{1.05}\\
\hline \hline
\hspace{-1mm}\makebox[\blob][r]{ } Spade (\textit{Ours}) & 
\textbf{57.55} \hspace{-2pt}\sdtext{1.22} & \textbf{72.84} \hspace{-2pt}\sdtext{0.97} & \underline{80.03} \hspace{-2pt}\sdtext{0.67} & 
\textbf{68.07} \hspace{-2pt}\sdtext{2.09} & \textbf{87.90} \hspace{-2pt}\sdtext{0.84} & \textbf{90.29} \hspace{-2pt}\sdtext{0.43} & 
\textbf{74.97} \hspace{-2pt}\sdtext{1.62} & \textbf{79.04} \hspace{-2pt}\sdtext{1.08} & \textbf{87.22} \hspace{-2pt}\sdtext{0.56}\\ [-0.2ex]
\hline
\end{tblr}
}
\vspace*{-6mm}
\end{center}
\end{table*}


\noindent\subsubsec{Local Feature Learning} \label{method2b}
Directly extending the proposed global strategies to the voxel level would be ineffective for three reasons.
1) Decoder outputs have higher resolutions, finer localized details, and require spatial equivariance rather than invariance.
2) We incur larger computational burdens when processing 3D feature maps at full resolution with contrastive projection heads which commonly contain more parameters than the backbone \cite{Chen2020ImprovedBW}.
3) Spatial alignments are not accurate enough to reliably assign local positives and negatives.

Instead of relying on full-resolution features, we use local features from the overlapping regions of crop pairs.
Given overlapping crops $P_a^j, P_a^k$ from image $I_a$ where $IoU(P_a^j, P_a^k) \geq o$ , we sample corresponding overlapping crops from other images (e.g., $P_b^j, P_b^k$).
We introduce \textit{three improvements to local feature learning}.
First, only logits in the overlapping regions (e.g., $Z_a^{j \cap k}, Z_b^{j \cap k}$) are used for representation learning.
This strategy mitigates the deleterious effects of misalignment errors while still promoting local understanding, retaining sample diversity, and limiting computation cost.
Second, we reverse the spatial transforms on logits (e.g., $Z'^{j \cap k}_a$=$\mathcal{T}_s^{-1}(Z_a^{j \cap k})$) before obtaining the final embeddings (e.g., $v'^{j \cap k}_a$=$h_l(Z'^{j \cap k}_a)$) so that representations are contrasted in their original orientations and equivariance is maintained. 
Finally, spatial relations in embeddings are preserved with 1x1 convolutions in $h_l$ instead of an MLP.
Note that $h_l$ also resizes logits to the local embedding size of $v_l \in \mathbb{R}^{C^{emb} \times H^{emb} \times W^{emb}}$.


In \textbf{local strategy 1} (\textit{L1}), we sample positives from the overlapping regions of the same crop pair and treat all $v_l \in M_l$ as negatives.
Thus, we have $\Lambda^-_{\textit{L1}}$=$M_l$, $\Lambda^+_{\textit{L1}}$=$\{v'^{j \cap k}_a, v'^{k \cap j}_a\}$.
Note that $v_a^{j \cap k}$ is the overlapping region of $v_a^j$ \& $v_a^k$ within $v_a^j$, while $v_a^{k \cap j}$ is the same corresponding area except in $v_a^k$.
\textbf{Local strategy 2} (\textit{L2}) debiases negatives that are above an overlap threshold (the same $o$ as previously defined).
More specificially, $\Lambda^-_{\textit{L2}}$=$\{v_i \ | \ IoU(v_{i \rightarrow t}, v_{a \rightarrow t}) \leq o, \forall v_i \in M_l\}$, $\Lambda^+_{\textit{L2}}$=$\{v'^{j \cap k}_a, v'^{k \cap j}_a\}$.
\textbf{Local strategy 3} (\textit{L3}) selects other corresponding overlapped regions as positives in addition to applying \textit{L2}: $\Lambda^-_{\textit{L3}}$=$\{v_i \ | \ IoU(v_{i \rightarrow t}, v_{a \rightarrow t}) \leq o, \forall v_i \in M_l\}$, $\Lambda^+_{\textit{L3}}$=$\{v'^{j \cap k}_a, v'^{k \cap j}_a, v'^{j \cap k}_b, v'^{k \cap j}_b, \ldots\}$.
Finally, \textbf{local strategy 4} (\textit{L4}) adopts the idea introduced in \textit{G3} that utilizes debiased negatives as positives:
$\Lambda^-_{\textit{L4}}$=$\{v_i \ | \ IoU(v_{i \rightarrow t}, v_{a \rightarrow t}) \leq o, \forall v_i \in M_l\}$, $\Lambda^+_{\textit{L4}}$=$\{v'^{j \cap k}_a, v'^{k \cap j}_a, v'^{j \cap k}_b, v'^{k \cap j}_b, \ldots\} \cup \{v_i \ | \ IoU(v_{i \rightarrow t}, v_{a \rightarrow t}) > o, \forall v_i \in M_l\}$.

\noindent\subsubsec{Reconstruction}
MG \cite{Zhou2019ModelsGG} elegantly fits into our pipeline by co-opting the data transformations used for contrasting views and complementing local representations via dense reconstruction.  
Without bells \& whistles, we directly reconstruct the spatially transformed version of the original crop (i.e. $\mathcal{T}_s(P)$).
The \textbf{reconstruction loss} is defined as:
\vspace*{-1mm}
\begin{equation}
   {\small{ \mathcal{L}^{MSE}_r = \left[ \mathcal{T}_s(P) - \sigma(g_{\theta} \circ f_{\theta}  \circ \mathcal{T}_i \circ \mathcal{T}_s(P) )\right]^2}}
\end{equation}
\vspace*{-5mm}

\noindent\subsubsec{Overall Loss \& Training Details}
Our \textbf{overall loss} $\mathcal{L}_{Spade}$ is:
\vspace*{-1mm}
\begin{equation}
{\small{ \mathcal{L}_{Spade} = \lambda \mathcal{L}^{CON}_g + (1 - \lambda) \mathcal{L}^{CON}_l  + \lambda_r \mathcal{L}^{MSE}_r}}
\end{equation}
\vspace*{-5mm}

\noindent where $\lambda$ and $\lambda_r$ are loss-weighing terms for contrastive learning and reconstruction, respectively.
For $\mathcal{L}_g^{CON}$ \& $\mathcal{L}_l^{CON}$, $\Lambda^{-/+}$ is populated based on the global and local strategies selected.
Note that local sets $\Lambda_l$ contain embeddings with spatial dimensions rather than vectors, so dot-products in Eq. \ref{nce} are spatially equivariant.
After weights are updated in an optimization step, we adjust the momentum parameters, $\epsilon$, from the regular model parameters, $\theta$, via $\epsilon \leftarrow \beta \epsilon + (1 - \beta) \theta$.


To compute $\mathbb{T}$ for volume alignment, we first crop out the background by thresholding at $-350$ Hounsfield Units, and downsample all images by a factor of 2.
Affine registration is performed using the negative normalized cross correlation metric and tri-linear interpolation. 
Optimization is run until convergence (minimum 50 iterations) with a 0.5 learning rate.

For network components, $f$ is a 3D Res2Net-50 (scale=4, stride=[1,2,2] in the first downsampling layer) \cite{gao2019res2net}, while $g$ is a lightweight decoder.
The global projection module $h_g$ follows \cite{Chen2020ImprovedBW} with average pooling to size 1x1x1, flattening, linear projection to 2048 channels, ReLU, and linear projection to embeddings of size $C_g^{emb}=128$.
The local projection module resizes features to 1x3x3 via average pooling, applies 1x1x1 convolution with 1024 filters, activates with ReLU, and outputs embeddings after a 1x1x1 convolution with 64 filters ($v_l \in \mathbb{R}^{64 \times 3 \times 3}, C_l^{emb}$=$64, H_l^{emb}$=$3, W_l^{emb}$=$3$; note the depth dimension is squeezed). 
In contrastive pretraining, we use $\mathcal{Q}_g$=$16000$, $\mathcal{Q}_l$=$1000$, $o$=$0.2$, $p$=$2$, $n^+$=$4$, and $\beta$=$0.99$. 
For losses, we set $\tau$=$0.2$, $\lambda$=$0.5$, and $\lambda_r$=$10$.

\vspace*{-1mm}
\section{Experiments} \label{exp}

\subsection{Pretraining Data and Details} \label{exp1}

\subsubsec{Data Description}
We demonstrate the efficacy our approach with torso CTs from two datasets as pretraining data since they present diverse anatomical structures, diseases, diagnostic tasks, and imaging settings.

\textbf{AMOS} \cite{ji2022amos} (CT \textbf{A}bdominal \textbf{M}ulti-\textbf{O}rgan \textbf{S}egmentation) provides 240 training and 120 test abdominal CTs from diverse sources.
We remove all ``abnormal images" with anisotropic spacing or sizing along the right/left and anterior/inferior axes.
Pretraining uses the remaining 200 training scans and is validated on 100 testing scans. 
Note that although masks are given for 15 organs, we do not use them in any capacity. 

\textbf{LUNA} \cite{Setio2016LUNAPulmonaryND} (CT \textbf{Lu}ng \textbf{N}odule \textbf{A}nalysis) consists of 888 chest CTs across 10 folds. 
The first 7 folds (623 images) and the remaining 3 folds (265 images) are used for pretraining and validation, respectively. 

\subsubsec{Implementation Details}
To \textbf{preprocess} all 823 pretraining \& 365 validation CTs, all images are resampled to 2$\times$0.7$\times$0.7 (mm) superior, anterior, right spacings, respectively (the median spacing in all pretraining datasets).
Intensities are clipped from $-1000$ to $1000$, and normalized between $0$ \& $1$.
Crops are obtained with scaling factor [0.5, 2], but ultimately resized to [32, 64, 64]; crops with mainly air are discarded.
We \textbf{train} using SGD (l.r. $0.0075$, mom. $0.9$) with cosine annealing \& batch size 24 for 500 epochs on a single V100 with PyTorch.

\begin{table}
\begin{center}
\caption{\label{tab:abl1}\textbf{Global feature learning approaches}. 
Pretrained on 5.2M AMOS \& LUNA crops, finetuned on 10\% of BCV labels (3 volumes), and evaluated on the test set.
}
\vspace*{-2mm}
\resizebox{0.8\columnwidth}{!}{
\begin{tblr}{
    colspec={l c l c || l},
    row{1} = {l,gray!50!black!15},
    row{2,6,12} = {gray!20!black!5}
    }
\hline 
\# & & Feature Init.  & \hspace*{2pt}$\mathcal{Q}_g$ & Dice (\%) \\ [-0.6ex] 
\hline \hline
\multicolumn{5}{l}{\textit{No Spatial Priors}} \\ [-0.6ex] 
\hline
1 & & Random  & - & 52.72  \\ [-0.8ex]
2 & \ssquare & PGL & - & 54.87 \ptext{+2.2}  \\ [-0.8ex]
3 & \ssquare & MoCo & 4k & 55.11 \ptext{+2.4}  \\ [-0.4ex]
\hline 
\multicolumn{5}{l}{\textit{Global Debiasing Strategies}} \\ [-0.6ex] 
\hline
4 & \dsquare & \textit{G1} ($n^+=4$)        & 4k & 55.49 \ptext{+2.8} \\ [-0.8ex]
5 & \dsquare & \textit{G2} ($n^+=4, o=0.0$) & 4k & 55.24 \ptext{+2.5} \\ [-0.8ex]
6 & \dsquare & \textit{G2} ($n^+=4, o=0.2$) & 4k & 56.37 \ptext{+3.7} \\ [-0.8ex]
7 & \dsquare & \textit{G2} ($n^+=4, o=0.4$) & 4k & 56.17 \ptext{+3.5} \\ [-0.8ex]
8 & \dsquare & \textit{G3} ($n^+=4, o=0.2$) & 4k & 56.58 \color{green!60!black!90} \textbf{(+3.9)} \color{black} \\ [-0.2ex]
\hline 
\multicolumn{5}{l}{\textit{Effect of Queue Size \& Debiasing}} \\ [-0.4ex] 
\hline
9 & \ssquare & MoCo                                & 1k  & 55.64 \ptext{+2.9}  \\ [-0.8ex]
10 & \ssquare & MoCo                               & 4k  & 55.11 \ptext{+2.4}  \\ [-0.8ex]
11 & \ssquare & MoCo                               & 16k & 54.98 \ptext{+2.3}  \\ [-0.8ex]
12 & \dsquare & \textit{G3} ($n^+=4, o=0.2$) & 1k  & 56.26 \ptext{+3.5} \\ [-0.8ex]
13 & \dsquare & \textit{G3} ($n^+=4, o=0.2$) & 4k  & 56.58 \ptext{+3.9} \\ [-0.8ex]
14 & \dsquare & \textit{G3} ($n^+=4, o=0.2$) & 16k & 56.83 \color{green!60!black!90} \textbf{(+4.1)} \color{black} \\ [-0.2ex]
\hline
\end{tblr}  
}
\vspace*{-8mm}
\end{center}
\end{table}

\vspace*{-1mm}
\subsection{Fine-tuning Data and Details} \label{exp2}

\subsubsec{Data Description} 
To gauge the quality of learned features, we specifically selected three tasks covering different torso regions, organs, modalities, and scales that were separate from the pretraining data but contain formerly seen structures.
For all datasets, we employ a 7:1:2 (train, validation, test) split and fine-tune using 10\%, 25\%, and 50\% of training images (if the number of training volumes isn't evenly divisible, we apply the floor function).
Following \cite{Zhou2021PreservationalLI}\cite{chaitanya2020contrastive}\cite{tao2020revisiting}, we use the class-averaged dice score for all segmentation evaluations.

\textbf{BCV} \cite{bcv2015} (CT \textbf{B}eyond the \textbf{C}ranial \textbf{V}ault) contains 30 abdominal CTs with 13 anatomical annotations.
The dataset is split into 21 training, 3 validation, and 6 testing images. 

\textbf{MMWHS} \cite{zhuang2013challengesmmwhs} (CT \& MR \textbf{M}ulti-\textbf{M}odality \textbf{W}hole \textbf{H}eart \textbf{S}egmentation) presents 20 labeled CT and 20 labeled MR images with annotations for seven cardiac structures. 
We treat each modality as its own downstream task and evaluate them independently.
For both CT \& MR subsets, we split the labeled images into 14 training, 2 validation, and 4 testing.

\subsubsec{Implementation Details}
For \textbf{preprocessing}, we clip values between $-1000$ \& $1000$ for CTs and Z-normalize for MRs.
We apply nnU-Net \cite{isensee2021nnu} spatial preprocessing and squish intensities between $0$ \& $1$ to be consistent with pretraining.
Regarding \textbf{data}, we sample crops (ensuring 50\% have foreground) with scaling factor [0.75, 1.25] and resize to [32, 96, 96]. 
Crops are augmented with mirroring, blurring, intensity scaling, and gamma.
We \textbf{train} using AdamW (l.r. 0.001, w.d. 0.01) with cosine annealing \& batch size 8 for $\approx$50k iterations on a single Titan-Xp with PyTorch (run time $\approx$16 hours).
We find higher weight decay to benefit, especially with limited labels.

\vspace*{-1mm}
\subsection{Baselines} \label{exp3}
\vspace*{-1mm}

We select state-of-the-art medical pretraining baselines like PCRL \cite{Zhou2021PreservationalLI} TransVW \cite{zHaghighi2021TransferableVW}, Rubik++ \cite{tao2020revisiting}, PCL \cite{Zeng2021PositionalCL}, and others.
We choose PCL over \cite{chaitanya2020contrastive} for its proposed improvements regarding slices near partition borders.
We make comparisons as fair as possible with PCL (a 2D method) by matching the number of parameters in 2D Res2Net with our 3D model via width increases.
FG \cite{dippel2021towardsfine} is omitted since global contrast coupled with reconstruction is explored by PCRL \cite{Zhou2021PreservationalLI} and by us in \autoref{res1}.
We exclude HSSL \cite{Zheng2021HierarchicalSL} for similar reasons in addition to the fact that it only uses 2D slices.   



\begin{table}
\begin{center}
\caption{\label{tab:abl2}\textbf{Local feature learning strategies \& Template choice.} 
Pretrained on 5.2M AMOS \& LUNA crops, finetuned on 10\% of BCV labels (3 volumes), and evaluated on the test set.}
\vspace*{-5mm}
\resizebox{\columnwidth}{!}{
\begin{tblr}{
    colspec={l l c || l},
    row{1} = {l,gray!50!black!15},
    row{2,10} = {gray!20!black!5}
    }
\hline 
 &  Feature Init. & $\mathcal{Q}_l $ & Dice (\%) \\ [-0.6ex] 
\hline 
\multicolumn{5}{l}{\textit{Local Debiasing Strategies}} \\ [-0.6ex] 
\hline
1 &  Ours$_{\text{global}}$ (\textit{G3}, $n^+=4, o=0.2, \mathcal{Q}_g=16$k)                      & - & 56.83   \\ [-0.8ex]
2 &  Ours$_{\text{global}}$ + Reconstruction    & - & 56.94 \ptext{+0.1} \\ [-0.8ex]
3 &  Ours$_{\text{global}}$ + \textit{L1} ($n^+=4, o=0.2$)  & 1k & 57.18 \ptext{+0.4}      \\ [-0.8ex]
4 &  Ours$_{\text{global}}$ + \textit{L2} ($n^+=4, o=0.2$)  & 1k & 57.41     \ptext{+0.6}       \\ [-0.8ex]
5 &  Ours$_{\text{global}}$ + \textit{L3} ($n^+=4, o=0.2$)  & 1k & 56.54     \ntext{-0.3}  \\ [-0.8ex]
6 &  Ours$_{\text{global}}$ + \textit{L4} ($n^+=4, o=0.2$)  & 1k & 56.13     \ntext{-0.7}  \\ [-0.8ex]
7 &  Ours$_{\text{proposed}}$ (\textit{G3} + \textit{L2} + Reconstruction)  & 1k & 57.55     \ptext{+0.7}       \\ [-0.2ex]
\hline 
\multicolumn{5}{l}{\textit{Template Selection}} \\ [-0.6ex] 
\hline
8 &  Ours$_{\text{proposed}}$                 & 1k & 57.55         \\ [-0.8ex]
9 &  Ours$_{\text{proposed}}$ (AMOS Template) & 1k & 57.66        \\ [-0.8ex]
10 &  Ours$_{\text{proposed}}$ (BCV Template) & 1k & 57.26         \\ [-0.8ex]
& \textit{Average} &  & 57.49  \\[-0.2ex]
\hline
\end{tblr}
}
\vspace*{-8mm}
\end{center}
\end{table}

\makeatletter
\newcommand{\raisemath}[1]{\mathpalette{\raisem@th{#1}}}
\newcommand{\raisem@th}[3]{\raisebox{#1}{$#2#3$}}
\makeatother

\section{Results and Discussion} \label{res}

\subsection{Component Studies} \label{res1}

\subsubsec{Global Feature Learning}
We first investigate the effectiveness of our global sampling strategies (see Table \ref{tab:abl1}; note \ssquare \hspace{0.6mm} \& \dsquare \hspace{1mm} indicate same-image positives \& inter-image positives, respectively).
For fair assessment, we train each method with 5.2M sampled patches and make queue sizes comparable for MoCo.
After pretraining, a randomly-initialized decoder is attached to the encoder \& the entire network is finetuned.

First, we note that incorporating negatives, even without debiasing, improves over positive-only approaches (see rows 2 \& 3).
Also, enriching positives with spatially corresponding patches (\textit{G1}, row 4) adds additional benefits. 
Debiasing negatives (rows 5-8), however, yields the largest increases when the overlap threshold $o$ is properly selected.
A threshold that's too low (row 5) may cause the undesirable removal of hard negatives (performing even worse than no debiasing in row 4). 
On the other hand, setting $o$ too high (row 7) prevents removal of damaging false negatives.
Useful to contrastive learning, $o$ can modulate the amount of mutual information that's distilled between samples (e.g., low $o$ amounts to negatives with low mutual information with the anchor).
This enables pretraining flexibility in the face of wide ranges of possible tasks \& data.

Next, we study the interactions between queue size and debiasing. 
In contrastive learning without debiasing (rows 9-11), we observe declines in performance as queue size increases.
This is may be attributed to the increasing number of false negatives; surprisingly, the decline is limited possibly because the vast majority of queue entries are valid negatives. 
In contrast, our debiasing strategies (rows 12-14) empower larger queue sizes where performance steadily improves. 
This affirms the principle that appropriate selections of both positives \& negatives facilitate contrastive representation learning.

\subsubsec{Local Feature Learning \& Template Choice}

In Table \ref{tab:abl2}, we explore the efficacy of local approaches by jointly pretraining using our best global strategy (\textit{G3} in row 1) and the indicated local strategy (note: the same $o$ \& $n^+$ are adopted from the global strategy).
To evaluate, we use the same precedure as experiments in Table \ref{tab:abl1}, except a single layer segmentation head (1x1x1 conv.) is appended to end of the decoder for pixel classification. 

We first compare our equivariant contrastive method (\textit{L1}, row 3) with a traditional reconstruction approach (row 2) and find the contrastive approach to be superior.
However, we integrate reconstruction since we observe performance improvements, and also qualitatively see acceleration in pretraining convergence \& speculate that it expedites initialization of sensible features which shortens feature warmup for contrastive tasks.
Next, we find that local strategies (rows 3-6) behave in stark contrast to global ones in that diversifying positives (rows 5-6) impairs features.
This is probably from rough correspondences yielding erroneous positives \& negatives, thus, blunting both cohorts.
So, our final proposed method utilizes \textit{G3}, \textit{L2}, and reconstruction. 
Comparing the training times of Spade to PCRL \cite{Zhou2021PreservationalLI}, our method takes 25.4 hours for 5.2 million patches while PCRL requires 28 hours. 

To study the sensitivity of our approach to different templates, we select 3 full-torso CTs: two from within the pretraining set (in AMOS), and one outside of pretraining data (in BCV). 
The only selection criteria is that the entire torso is covered.
From Table \ref{tab:abl2}, we conclude that although there's a slight reduction in performance for the BCV template, Spade performs reasonably well for all three.
This affirms the notion that approximate spatial correspondences are sufficiently effective for semantic comparisons. 
%

\vspace*{-1mm}
\subsection{Comparison with State-of-the-Arts} \label{res2}
\vspace*{-1mm}

From our main experiments in Table \ref{tab:main}, we make the following observations. 
\textbf{1)} Spade generally \textit{outperforms recent state-of-the-art medical pretraining methods}.
This supports our assumption that spatial priors are effective for predicting semantic similarity and transferring this knowledge into useful representations. 
This also shows that simple sampling strategies can beat fairly complex pipelines like \cite{Zhou2021PreservationalLI} which uses multiple regularization approaches like mixup, attention modules, and transformation embeddings. 
\textbf{2)} Our method \textit{benefits performance the most when annotations are more limited}.
This is a desirable property for pretraining since its primary objective is to improve representations when annotations are most scarce due to acquisition costs. 
However, this is a double-edged sword in that when there's more labels (e.g. MMWHS 50\% labels), the cost of pretraining may not out-weight the benefit.
\textbf{3)} We affirm that \textit{effective negative samples are central to representation learning} in 3D radiology images. 
Our studies on sampling strategies (see \autoref{res1}, \autoref{res2}) indicate that debiasing improves downstream performance more than enriching the diversity of positives. 
The task of making two patches with comparable semantics similar may be an easier task (courtesy of the curse of dimensionality) than contrasting similar patches. 
More studies regarding metric learning in medical images are needed to further explore this observation. 


\vspace*{-1mm}
\section{Conclusion} \label{concl}

In this work, we present a contrastive learning framework, Spade, that leverages spatial correspondences between 3D radiology images to enrich positive pairs and debias false negatives. 
Our studies indicate that Spade adds little additional computation, is highly effective in learning representations for downstream segmentation tasks, and hopefully spurs additional work in discovering more effective priors for contrastive learning with the abundant medical data we have at our disposal.

\bibliographystyle{plain}
\bibliography{refs}{}


\end{document}